\documentclass[letterpaper, 10 pt, conference]{ieeeconf}  

\IEEEoverridecommandlockouts                              

\usepackage{graphicx} 
\usepackage{mathptmx} 
\usepackage{amsmath} 
\usepackage{amssymb}  
\usepackage{bm}
\usepackage{amsmath,amssymb}
\usepackage{mathrsfs}
\usepackage{cite}
\usepackage{url}
\usepackage{subfigure}
\usepackage{threeparttable}

\title{\LARGE \bf
Chat-PM: A Class of Composite Hybrid Aerial/Terrestrial Precise Manipulator
}

\author{Yihang Ding, Xiaoyu Ji, Lixian Zhang, \emph{Fellow, IEEE}, Yifei Dong, Tong Wu, Chengzhe Han
\thanks{*This work was not supported by any organization. (\textit{Yihang Ding and Xiaoyu Ji are co-first authors.})(\textit{Corresponding author: Lixian Zhang.})}
\thanks{Yihang Ding, Xiaoyu Ji, Lixian Zhang, Yifei Dong, Tong Wu, and Chengzhe Han are with the School of Astronautics, Harbin Institute of	Technology, Harbin, 150080, China. 
	{\tt\small  yhding@stu.hit.edu.cn; xyji@stu.hit.edu.cn; lixianzhang@hit.edu.cn; yfdong@stu.hit.edu.cn; 	wu\_tong@hit.edu.cn;  czhan@stu.hit.edu.cn}}%
}

\begin{document}

\maketitle
\thispagestyle{empty}
\pagestyle{empty}

\begin{abstract}
This paper concentrates on the development of Chat-PM, a class of composite hybrid aerial/terrestrial manipulator, in concern with composite configuration design, dynamics modeling, motion control and force estimation. Compared with existing aerial or terrestrial mobile manipulators, Chat-PM demonstrates advantages in terms of reachability, energy efficiency and manipulation precision.  
To achieve precise manipulation in terrestrial mode, the dynamics is analyzed with consideration of surface contact, based on which a cascaded controller is designed with compensation for the interference force and torque from the arm. 
Benefiting from the kinematic constraints caused by the surface contact, the position deviation and the vehicle vibration are effectively decreased, resulting in higher control precision of the end gripper. For manipulation on surfaces with unknown inclination angles, the moving horizon estimation (MHE) is exploited to obtain the precise estimations of force and inclination angle, which are used in the control loop to compensate for the effect of the unknown surface. Real-world experiments are performed to evaluate the superiority of the developed manipulator and the proposed controllers.
\end{abstract}

\section{Introduction}
\begin{figure}[tbp]
	\centering
	\includegraphics[width=0.46\textwidth]{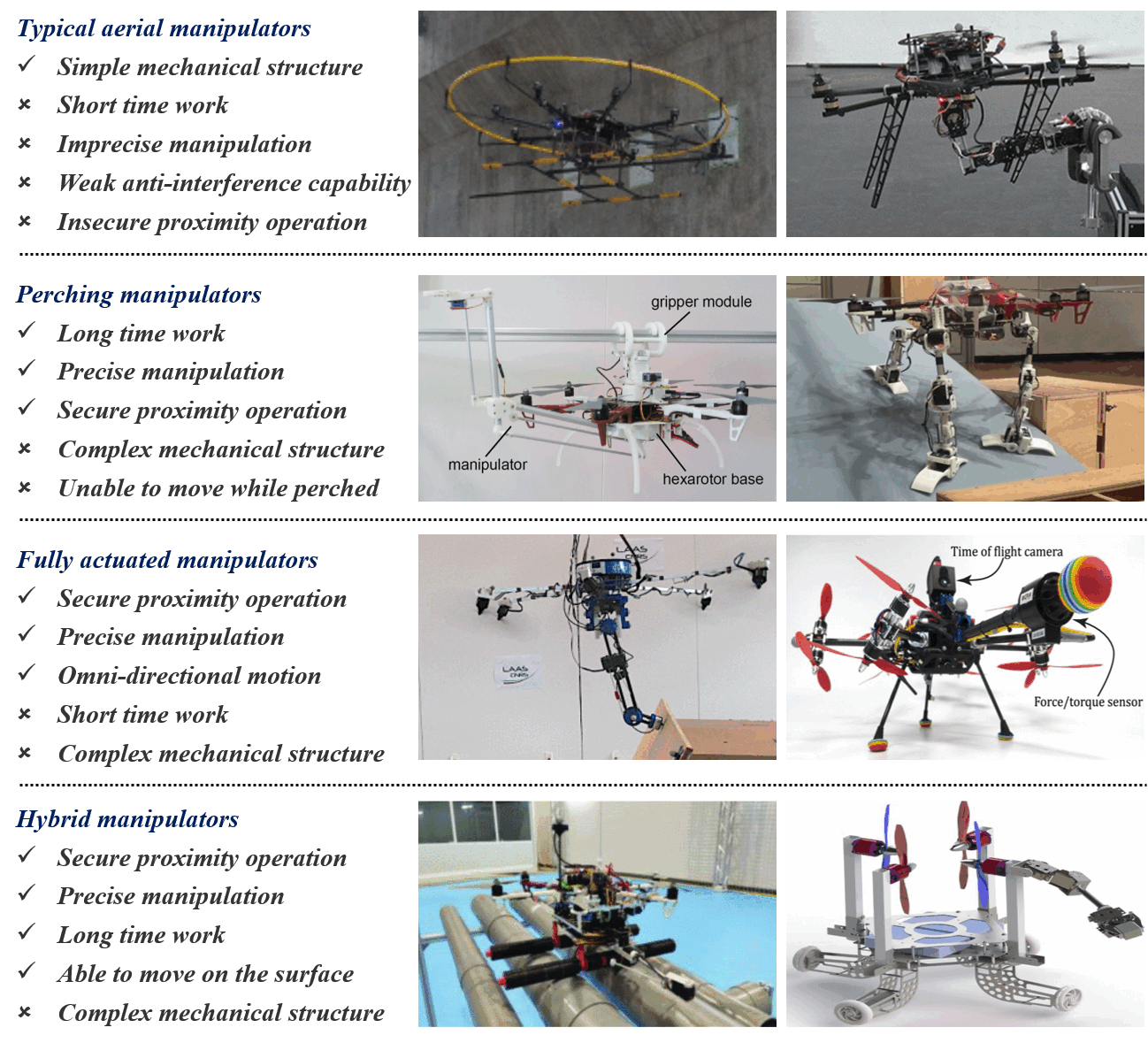}
	\caption{Summary of the manipulators, which are roughly divided into four categories: typical aerial ones, perched ones, fully actuated ones and hybrid ones, cf.\cite{8206398,7139968,8452699,8968529,8928943,9295362,9341639,6618146}.}
	\label{intro}
		\vspace{-0.5em}
\end{figure}

In recent years, Unmanned Aerial Manipulators (UAMs) have attracted considerable attention due to their grasping and carrying capabilities, which are sorely needed in high-altitude operation, fire rescue and other hazardous scenes\cite{6385917,thomas2014toward,8206398,7139968,9981106}.
Among the various kinds of aerial manipulators (shown in Fig. \ref{intro}), the rotor-driven ones show unique advantages in high flexibility and agility, thus widely used in various manipulation tasks \cite{8206398}. Nevertheless, the high energy consumption severely limits their working time and scope. Also, the disturbances caused by wind or movement of the robotic arm may lead to significant tracking errors.
To address the two issues, the aerial manipulators with perching ability \cite{8452699,8968529} and full-actuation mechanism \cite{8928943,9295362} are increasingly studied, but simultaneously achieving low energy consumption and high
control precision remains a dilemma.


In terms of energy consumption and control precision, recent efforts are devoted to the hybrid aerial/terrestrial manipulator, so as to expand the admissible workspace, reduce energy consumption, or ensure operation safety \cite{9341639,6618146,9517691}. 
In terrestrial mode, the utilization of ground support facilitates a significant reduction in rotor thrust, thereby enabling lower energy consumption.
Furthermore, by attaching to the surface, operation tasks can be executed stably and efficiently.
However, existing configurations rely on supplementary actuators such as motors or grippers \cite{8206398,9341639,6618146} for terrestrial locomotion, which increases the mechanical complexity and degrades the payload capacity and endurance.
Moreover, the complex mechanical structure often restricts the landing agility and operation capabilities, thus preventing the platform from dense environments and complex manipulations.
It is also worth noting that almost all the existing studies on hybrid aerial/terrestrial manipulators seldom consider the operation on surfaces with various inclination angles. 
The dynamics of the manipulator on inclined surfaces and ground are different in supporting forces and frictional forces. The mainstream control methods \cite{9341639,6618146} developed for the horizontal surface are rather inadequate to guarantee the accuracy of operation and motion on inclined surfaces. 
Therefore, it is significant to explore new mechanical designs and corresponding control approaches for hybrid aerial/terrestrial manipulators to perform precise operation tasks on surfaces usually with unknown inclination angles.

In this paper, we present a new hybrid manipulator named {\bf{Chat-PM}} ({\textbf{C}}omposite {\bf{H}}ybrid {\bf{A}}erial/{\bf{T}}errestrial {\bf{P}}recise {\bf{M}}anipulator) to overcome the configuration limitations. Chat-PM is designed to be collision-tolerant, allowing for safe landing, moving and grasping objects on inclined surfaces or walls.
The main contributions of this work include:
\textit{(i)} A new composite configuration of hybrid aerial/terrestrial manipulators is presented, where the passive wheels are adopted to enable terrestrial locomotion with low weight and structural complexity.
\begin{figure}[tbp]
	\centering
	\includegraphics[width=0.47\textwidth]{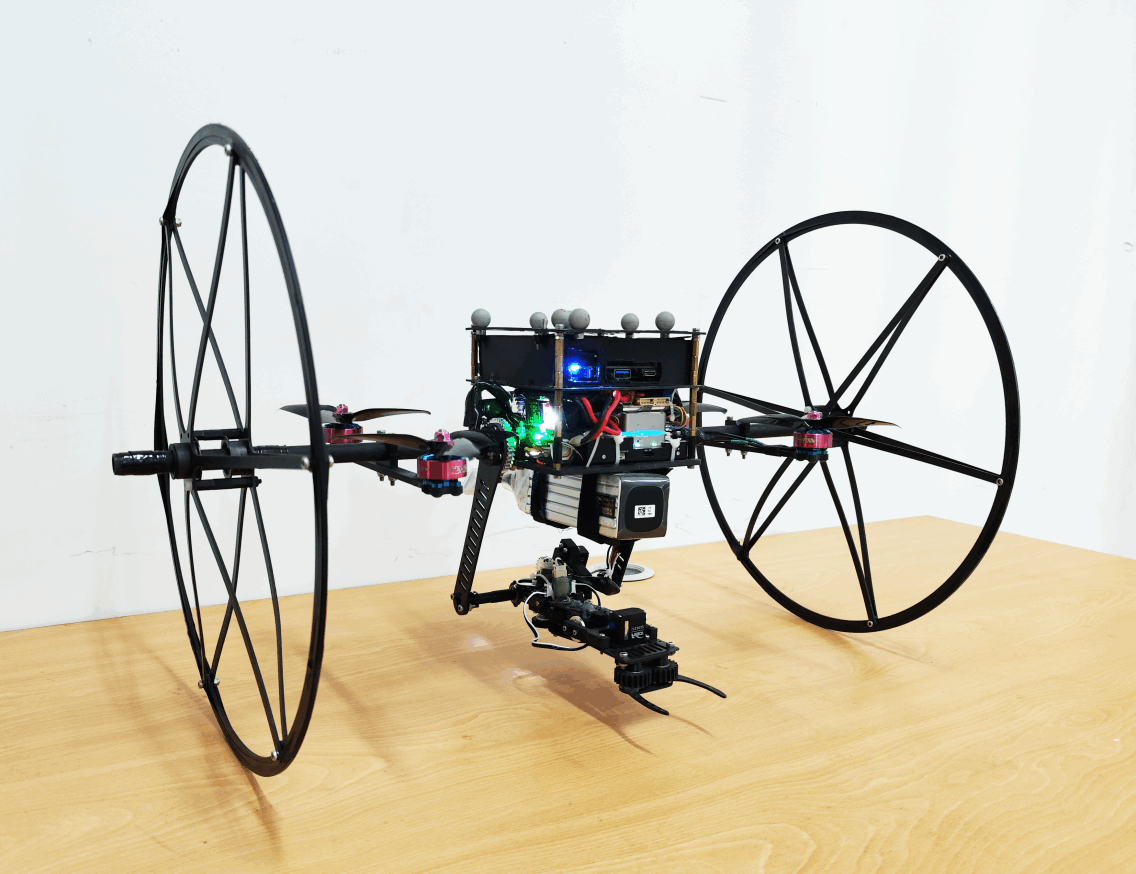}
	\caption{The prototype of Chat-PM, equipped with two independent passive wheels and four degrees of freedom robotic arm.}
	\label{prototype}
\end{figure}
\textit{(ii)} A unified control approach for Chat-PM on surfaces with various inclination angles is developed, and the precision of control and operation is significantly improved by attaching to the surface.
\textit{(iii)} The methodology of moving horizon estimation (MHE) is utilized to estimate the contact force and surface inclination. Based on the obtained angle estimation, the controller parameters can be scheduled to improve the tracking performance on surfaces with unknown inclinations.
The remainder of the paper is organized as follows: Section II describes the mechanical design. Section III covers the dynamics model and Section IV focuses on the multimodal control and force estimation. The experimental results are presented in Section V and Section VI concludes this letter.

\section{Mechanical Design}
In this section, the mechanical design and hardware integration of Chat-PM are presented, with a prototype shown in Fig. \ref{prototype}. To attain the collision-tolerant ability with lightweight, carbon fiber is adopted as the material for the main structure of Chat-PM, with some connection parts made of Thermoplastic Urethane (TPU) plastic prints for impact toughness. The main chassis of Chat-PM is a quadrotor with a diagonal wheelbase of $330 mm$, where four GTS2207 brushless DC motors (1860KV) equipped with 5-inch three-blade propellers provide thrust and torque. The Intel NUC onboard computer executes the software modules, including motion control of the robotic arm and position control of the vehicle. Meanwhile, the CUAV V5+ autopilot\footnote{https://doc.cuav.net/flight-controller/v5-autopilot}, with the PX4 firmware\footnote{https://github.com/PX4/PX4-Autopilot}, is used for attitude/thrust control and providing IMU measurements. For high-precision state estimation, an EKF is implemented on the NUC to fuse the information from IMU and OptiTrack system, which can significantly benefit position control precision in indoor scenes. The overall size of the prototype is $480\times340\times340mm$ and the weight is 2.4kg, the lightweight robotic arm is about 180g.

For terrestrial locomotion, the equipped passive wheels enable a series of motions on surfaces. The spoke structure can absorb more impact force while reducing the weight and the interference from the airflow. In addition, the wheels can protect the propellers from a collision with the surface or some obstacles.

\begin{figure}[tbp]
	\centering
	\includegraphics[width=0.5\textwidth]{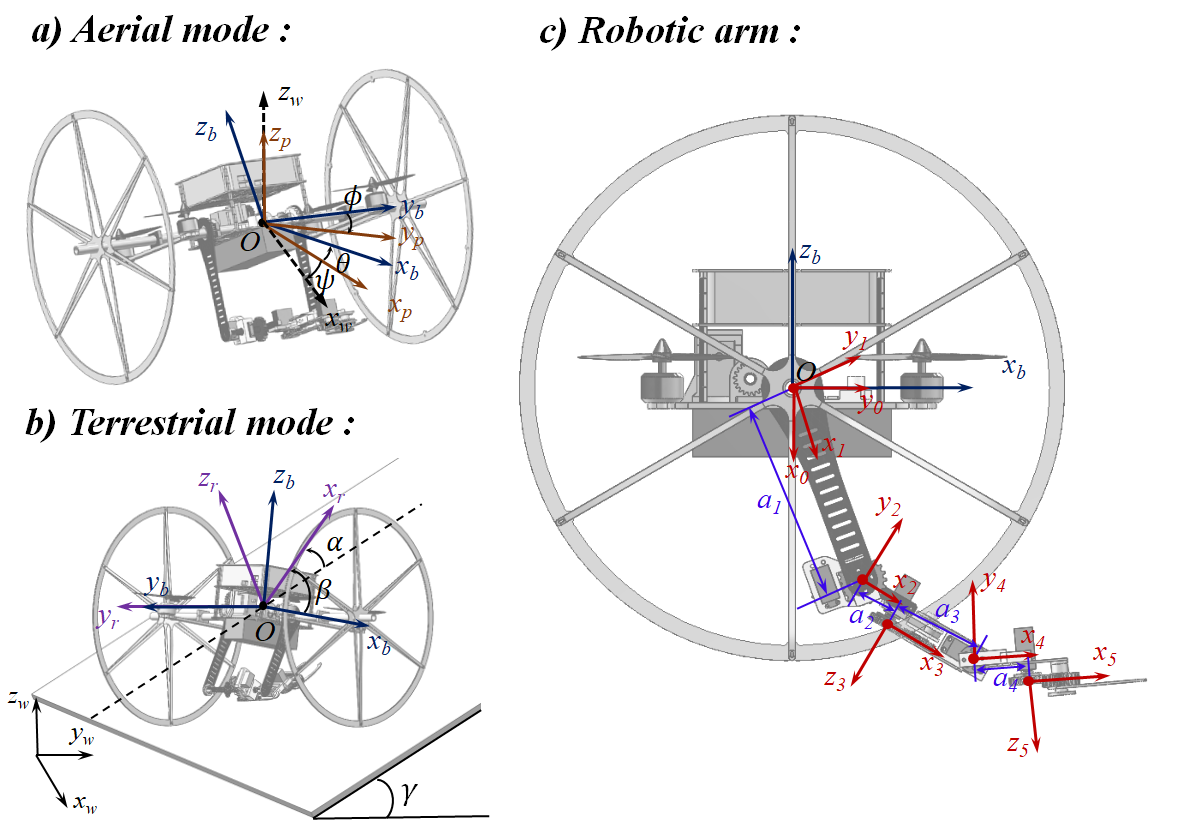}
	\caption{The coordinate frames for Chat-PM, consisting of aerial mode, terrestrial mode and robotic arm, in which $\bm{w}$, $\bm{b}$ and $\bm{r}$ denote the world, body and rolling frames, respectively.}
	\label{coordinate_frames}
\end{figure}

Before ending this section, the 4-DOF robotic arm is of lightweight design with the structure of the links and joints as shown in Fig. \ref{coordinate_frames}. The first link is mounted symmetrically on both sides of the body and hence can rotate unhindered all-directionally. The symmetrically mounted design ensures that the reaction force/torque of the arm is applied to the center of the vehicle, reducing interference caused by arm movement. The third joint is designed to swing around the rotation axis $\bm{\hat{z}_3} $, significantly enlarging the admissible working range. Each joint is driven by a digital servomotor with a magnetic encoder, which can provide angle and angular rate feedback. All servomotors are connected to a USB-to-USART interface of the onboard computer via the TTL serial bus so that the angle commands can be transmitted with a high communication rate and high reliability, satisfying the real-time demand of motion control for mobile operations.

\section{Dynamics Model}
This section is devoted to the dynamics modeling, including the bi-modal vehicle dynamics and the robotic arm dynamics. 
Also, the interactions between the vehicle and the arm and nonholonomic constraints are analyzed. 
The used coordinate frames are depicted in Fig. \ref{coordinate_frames}, where $\bm{w}$ denotes the world frame with $\bm{\hat{z}_w} $ pointing in the opposite direction of the gravity vector; $\bm{b}$ a body-fixed frame with $\bm{\hat{z}_b} $ normal to the propellers; $\bm{p}$ the frame with $\bm{\hat{z}_p}$ parallel to $\bm{\hat{z}_w}$;
$\bm{r}$ the rolling frame with $\bm{\hat{z}_r} $ perpendicular to the inclined surfaces (also including the horizontal ones) and $\bm{\hat{y}_r} $ parallel to $\bm{\hat{y}_b} $; $\bm{S_i} ({i}=0,1,2,3,4,5)$ denotes the joint coordinate system of the robotic arm. The position and attitude of the body frame w.r.t. the world frame are denoted as $\bm{r}=[x,y,z]^T$ and  $\bm{\eta}=[\phi,\theta,\psi]^T$, respectively; $\bm{\omega}=[p,q,r]^T$ the angular rate in the body frame; ${R}_b^w$ the rotation matrix of the body frame w.r.t. the world frame; ${R}_0^b$ the rotation matrix between the base frame $\bm{S_0}$ of the arm and the body frame.
\begin{table}[tbp]
	\caption{D-H Parameters}
	\label{DH-table}
	\vspace{-1.em}
	\begin{center}
		\begin{tabular}{c c c c c}
			\hline
			$i $   & $\alpha_{i-1}$(rad) 	& $a_{i-1}$(m) 	& $d_i$(m) & $\theta_i$(rad) \\
			\hline
			1    & 0   				& 0		&0&$\theta_1$ \\
			2   & 0        		& $0.134$ &0&$\theta_2$	 \\
			3   &	$-\pi/2$   	    	& 	$0.028$ &0.013&$\theta_3$		\\
			4   & $+\pi/2$&$0.038$&0&$\theta_4$ \\
			5   & $-\pi/2$&$0.034$&0.015&$\theta_5$ \\
			\hline
		\end{tabular}
	\end{center}
	\vspace{-1.5em}
\end{table}
\subsection{Aerial Mode}
By Newton-Euler equations, the translational and rotational dynamics of Chat-PM in aerial mode can be formulated as
\begin{equation}
\left\{\begin{array}{l}
m \ddot{\boldsymbol{r}}+m \boldsymbol{g}={R}_b^w(\boldsymbol{F}+\boldsymbol{f}) \\
{I} \dot{\boldsymbol{\omega}}+\boldsymbol{\omega} \times {I} \boldsymbol{\omega}=\bm{M}+\boldsymbol{\tau}
\end{array}\right.
\end{equation}
where $m$ denotes the mass of the vehicle chassis excluding the robotic arm; $\ddot{\bm{r}}\in\mathbb{R}^3$ denotes the linear acceleration of the body frame w.r.t. the world frame; $\bm{g}=[0,0,g]^T$ denotes the gravity vector; ${I}\in\mathbb{R}^{3\times3}$ denotes the inertial matrix that varies with the motion of the arm, which can be calculated using the Parallel-Axis Theorem \cite{abdulghany2017generalization}; 
$\bm{F}\in\mathbb{R}^3$ and $\bm{M}\in\mathbb{R}^3$ represent the force and torque acting on the center of the vehicle by rotors, respectively; $\bm{f}\in\mathbb{R}^3$ and $\bm{\tau}\in\mathbb{R}^3$ represent the force and torque disturbances on the vehicle, respectively. For Chat-PM, these disturbances are mainly caused by the movement of the robotic arm.

\subsection{Terrestrial Mode}
While rolling on surfaces, the Chat-PM is not only affected by the contact forces but also subject to additional kinematic constraints. 
As shown in Fig. \ref{coordinate_frames}, consider that the Chat-PM is moving on a surface with inclination $\gamma$. Then, the heading of the vehicle along the surface can be described by $\alpha$, which is the angle around the axis ${\hat{\bm{z}}}_{\bm{r}}$.
To ensure full contact between the wheels and the surface,  the roll angle $\phi$ and the yaw angle increment $\Delta\psi $ should be constrained by
\begin{subequations}
	\begin{equation}
		\phi=\arctan \ \frac{-\sin \gamma \sin \alpha}{\sqrt{\cos ^2 \alpha+\sin ^2 \alpha \cos ^2 \gamma}}
	\end{equation}
	\begin{equation}
		\Delta \psi=\arctan (\cos \gamma \tan \alpha)
	\end{equation}
\end{subequations}
In specific, the constraints will degenerate into $\Delta\psi = \alpha$ or $ \phi=-\alpha$ in the special case of moving on the horizontal ground or vertical wall, respectively.
Define the angle between ${\hat{\bm{x}}}_{\bm{r}}$ and the horizontal surface as the inclination angle $\delta$ of the surface as follows
\begin{equation} \delta = \arctan \ \frac{\sin\gamma \cos\alpha}{\sqrt{\sin^2\alpha + \cos^2\alpha \cos^2 \gamma}}\end{equation}
Based on the aforesaid solved angles, the rotation matrix between coordinate frames can be obtained. ${R}^r_w$ the rotation matrix of the world frame w.r.t. the rolling frame is denoted as
\begin{equation} {R}^r_w = {R}_z(-\alpha) {R}_y(\gamma) {R}_z(\Delta\psi-\psi) \end{equation}
Similar to ${R}^r_w$, the rotation matrix of the body frame w.r.t. the rolling frame is 
\begin{equation} {R}^r_b = {R}_y(\delta+\theta) 
 \end{equation}
where $\delta$ and $\theta$ are represented in Fig. \ref{coordinate_frames}. 
Then, note that the force ${F_r}$ and torque ${M_r}$ of the rotors in the rolling frame can be obtained as
\begin{subequations}
	\begin{equation}
		\bm{F}_r={R}_b^r \bm{F}=\left[\begin{array}{c}
			U_1 sin{(\delta+\theta)} \\
			0 \\
			U_1 cos{(\delta+\theta)}
		\end{array}\right] 
	\end{equation}
	\begin{equation}
		\bm{M}_r={R}_b^r \bm{M}=\left[\begin{array}{c}
			U_2 cos{(\delta+\theta)}+U_4 sin{(\delta+\theta)} \\
			U_3 \\
			-U_2 sin{(\delta+\theta)}+U_{4} cos{(\delta+\theta)}
		\end{array}\right]
	\end{equation}
\end{subequations}
The force and torque of the arm acting on the vehicle in the rolling frame, $^r\bm{f}_0$ and $^r\bm{n}_0$, are calculated by
\begin{subequations}
	\begin{equation}
		^r\bm{f}_0 ={{R}^r_b} {{R}^b_0} {^0\bm{f}_0} 
	\end{equation}
	\begin{equation}
		^r\bm{n}_0 ={{R}^r_b} {{R}^b_0} {^0\bm{n}_0}
	\end{equation} 
\end{subequations}
where ${^0\bm{f}_0}$ and ${^0\bm{n}_0}$, respectively, denote the force and torque acting on the vehicle by the base link of the arm, and more details can be found in the next subsection.
Then, the expression of the gravity of the vehicle in the rolling frame is
\begin{equation} \bm{f}_g = m {{R}^r_w} \bm{g}
 \end{equation}
Assuming that Chat-PM moves on the surface without lateral slip, the normal supporting force is expressed as $\bm{F}_n = \bm{F}_{left}+\bm{F}_{right}$. Given the rolling friction coefficient $\mu_r$ and the half length of the vehicle $l$, the rolling friction and the rolling resistance torque can be expressed as
\begin{subequations}
	\begin{equation}
		\bm{f}_{roll}=\bm{f_l+f_r}=\mu_r\left\|\bm{F}_{left}+\bm{F}_{right}\right\| \hat{\bm{x}}_{{r}} 
	\end{equation}
	\begin{equation}
		\bm{\tau}_{roll}=
		l\left\|\bm{f}_l - \bm{f}_r\right\| \hat{\bm{z}}_{{r}} 
	\end{equation}
\end{subequations}
Considering the inertial torque of wheels, the resistance force acting on the vehicle along the axis ${\hat{\bm{x}}}_{{r}}$ satisfies
\begin{equation} \bm{f}_{wh} = \frac{2I_{wh}\dot{v}_{rx}}{r^2_{wh}}{\hat{\bm{x}}_r} 
\end{equation}
where $r_{wh}$ and $I_{wh}$ denote the radius and the rotational inertia around the rotating axis of the wheel, respectively. Then, the resistance torque against the angular acceleration of the vehicle about ${\hat{\bm{z}}}_{{r}}$ is 
\begin{figure}[tbp]
	\centering
	\includegraphics[width=0.499\textwidth]{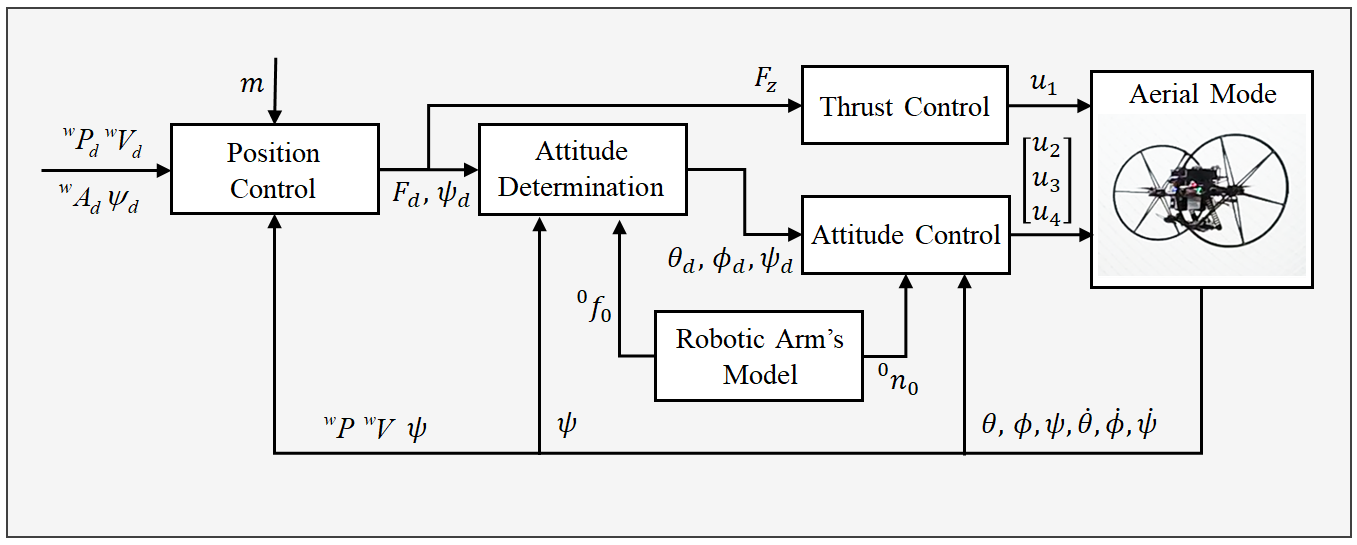}
	\caption{The architecture of aerial mode controller.}
	\label{aerial-controller}
\end{figure}
\begin{equation} \bm{\tau_{wh}} = \frac{2I_{wh} {\ddot{\alpha}}l^2}{r^2_{wh}}{\hat{\bm{z}}_r} 
\end{equation}
Finally, under the constraints of no-slip along ${\hat{\bm{y}}}_{{r}}$ and no rotation around ${\hat{\bm{x}}}_{{r}}$, the complete terrestrial dynamics of Chat-PM in the rolling frame can be formulated as follows
\begin{equation}
	\left\{\begin{array}{l}
		m \dot{v}_{r x}=\left(\bm{F}_r-\bm{f}_g-\bm{f}_{wh}-\bm{f}_{roll}-{ }^r \bm{f}_0\right) {\widehat{\bm{X}}_{\bm{r}}} \\
		\left(\bm{f}_g+{ }^r \bm{f}_0\right) {\widehat{\bm{Y}}_{\bm{r}}} \leq \mu_{s \operatorname{max}}\left\|\bm{F}_n\right\| \\
		\left(\bm{F}_n+\bm{F}_r-\bm{f}_g-{ }^r \bm{f}_0\right) {\widehat{\bm{Z}}_{\bm{r}}}=0 \\
		\left(\bm{M}_r-\bm{\tau}_g-{ }^r \bm{n}_0\right) \bm{\widehat{\bm{X}}_{\bm{r}}}=\left\|\bm{F}_L-\bm{F}_R\right\| l \\
		I_{y y}^r \ddot{\beta}=\left(\bm{M}_r-\bm{\tau}_g-{ }^r \bm{n}_0\right) {\widehat{\bm{Y}}_{\bm{r}}} \\
		I_{z z}^r \ddot{\alpha}=\left(\bm{M}_r-\bm{\tau}_{roll}-\bm{\tau}_{wh}-\bm{\tau}_g-{ }^r \bm{n}_0\right) {\widehat{\bm{Z}}_{\bm{r}}}
	\end{array}\right.
\end{equation}
where ${\widehat{\bm{X}}}_{\bm{r}}$, ${\widehat{\bm{Y}}}_{\bm{r}}$ and ${\widehat{\bm{Z}}}_{\bm{r}}$ denote the computed components along the three axes of the rolling frame, respectively; $\mu_{smax}$ is the maximum coefficient of sliding friction; $\bm{\tau}_g$ represents the torque induced by the offset of the center of mass; $\ddot{\beta}$ and $\ddot{\alpha}$ denote the angular accelerations about ${\hat{\bm{y}}}_{{r}}$ and ${\hat{\bm{z}}}_{{r}}$, respectively; $I_{yy}^r$ and $I_{zz}^r$ denote the inertial torque of the vehicle about ${\hat{\bm{y}}}_{{r}}$ and ${\hat{\bm{z}}}_{{r}}$, respectively.

\subsection{Modeling of 4-DOF Robotic Arm}
It is worth noting that the disturbances caused by the robotic arm's movement significantly affect both aerial and terrestrial modes dynamics.
To precisely evaluate the force and torque from the arm, it is necessary to model the dynamics of the robotic arm.
Based on the coordinate frames defined in Fig. \ref{coordinate_frames}, the Denavit-Hartenberg (D-H) method \cite{4252158,7081929} is used to describe the spatial relationship between each link from the base link to the end-effector. The obtained D-H parameters are shown in Tab. \ref{DH-table}. 
Therefore, the transformation matrix between the coordinate frames $S_5$ and $S_0$ can be constructed, denoted as
$
^0_5{T} = {}^0_1{T} {}^1_2{T} {}^2_3{T} {}^3_4{T} {}^4_5{T} 
$, where ${_i^{i-1}}{T}\in\mathbb{R}^{4\times4}$ denotes the transformation matrix between adjacent coordinate frames. 

According to the spatial relationship of the links, the forces and torques acting on the base coordinate frame can be calculated using the iterative Newton–Euler dynamic formulation \cite{7474862,126092}, given as follows
\begin{subequations}
\begin{align}
{ }^i \bm{F}_i&=m_i{ }^i \dot{\bm{v}}_{C_i} \\
{ }^i \bm{N}_i &={ }^{C_i} {I}_i{ }^i \dot{\bm{\omega}}_i+{ }^i \bm{\omega}_i \times{ }^{C_i} {I}_i{ }^i \bm{\omega}_i\\
{ }^i \bm{f}_i&={ }_{i+1}^i {R}^{i+1} \bm{f}_{i+1}+{ }^i \bm{F}_i \\
{ }^i \bm{n}_i&={ }^i \bm{N}_i+{ }_{i+1}^i {R}^{i+1} \bm{n}_{i+1}+{ }^i \bm{P}_{C_i} \times{ }^i \bm{F}_i \\
			&+{ }^i \bm{P}_{i+1} \times{ }_{i+1}^i {R}^{i+1} \bm{f}_{i+1}  \notag
\end{align} 
\end{subequations}
%
%
where $ ^i\bm{f}_j$ and $^i\bm{n}_j$ represent force and torque acting on link $i$ by link $j$, respectively; $ ^i\bm{F}_j$ and $ ^i\bm{N}_j$ express the inertial force and torque acting on the centroid of link $i$, respectively; $ ^i\bm{P}_{C_i}$ and $ ^i\bm{P}_j$ denote positions of the centroid of link $i$ and the origin of the frame $S_j$ w.r.t. the frame $S_i$, respectively; $m_i$ and $ ^{C_i}{I}_i$ represent the mass and the inertia matrix of link $i$, respectively; $^i\bm{\omega}_i$ is the angular rate of link $i$ w.r.t. the frame $S_i$. To compensate for the effect of the gravity on the links,  $^0{\dot{\bm{v}}}_0$ is set to be ${R}_b^0{R}_w^b(\ddot{\bm{r}}-\bm{g})$ in this letter here.

Note that the calculation of interference force and torque from robotic arm involves two processes. Firstly, the angular rate, angular acceleration, linear velocity, and linear acceleration of the centroid of each link are calculated by iterating outward from the $\bm{S_0}$ base frame.
Thereby, the inertial force and torque acting on each link can be calculated by (13a) and (13b). Moreover, the interaction force and torque acting on each link can be computed by (13c) and (13d) in an iterative inward form from the $\bm{S_5}$ end-effector frame. In particular, $ ^5\bm{f}_5$ and $ ^5\bm{n}_5$ are set equal to zero when the arm is not in contact with any object. 
Finally, the force and torque acting on the base link, $ ^0\bm{f}_0$ and $ ^0\bm{n}_0$, can be calculated, and introduced into both aerial and terrestrial dynamics.
\begin{figure}[tbp]
	\centering
	\includegraphics[width=0.499\textwidth]{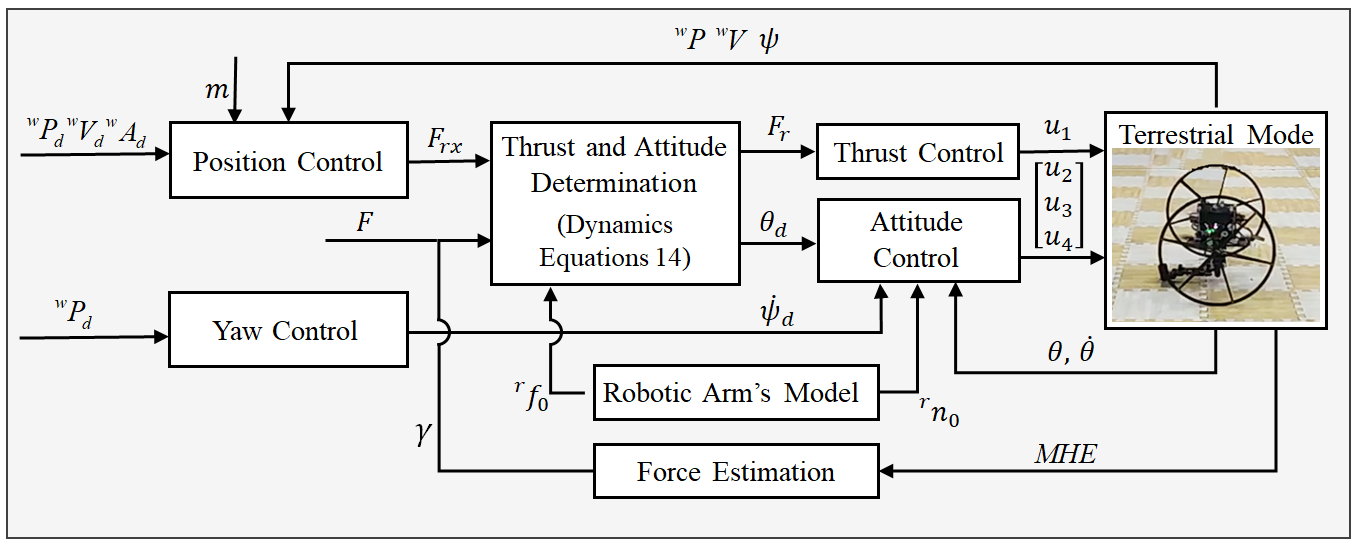}
	\caption{The architecture of terrestrial mode controller.}
	\label{terrestrial-mode controller}
\end{figure}
\section{Multimodal Control And Force Estimation}
This section elaborates on the controller designs of the vehicle for both aerial and terrestrial modes, and the latter includes the motion on surfaces with various inclinations. 
An estimator for the normal contact force between the vehicle chassis and the surface in terrestrial mode is developed to calculate the inclination of the unknown slope.
\subsection{Aerial-Mode Controller}
As illustrated in Fig. \ref{aerial-controller}, within the motion control loop for aerial mode, the desired state world coordinate frame ${{^w}P_d,{^w}V_d,{^w}A_d,\psi_d}$ from the high-level planner is transmitted to the aerial-mode controller.
Similar to the mainstream flight control methods studied for hybrid aerial/terrestrial vehicles,
the control strategy in this work consists of position control, attitude control, and throttle control. In particular, both the position and attitude control loops are equipped with cascaded controllers, with design details available in \cite{9832723,9691888,8968276,brescianini2013nonlinear}.
In addition, the interference force and torque caused by the robotic arm's motion, $^0\bm{f}_0$ and $^0\bm{n}_0$, can be calculated by (13) and thereby be compensated for in the control loop, such that the motion performance can be improved.
\begin{figure}[tbp]
	\centering
	\includegraphics[width=0.3\textwidth]{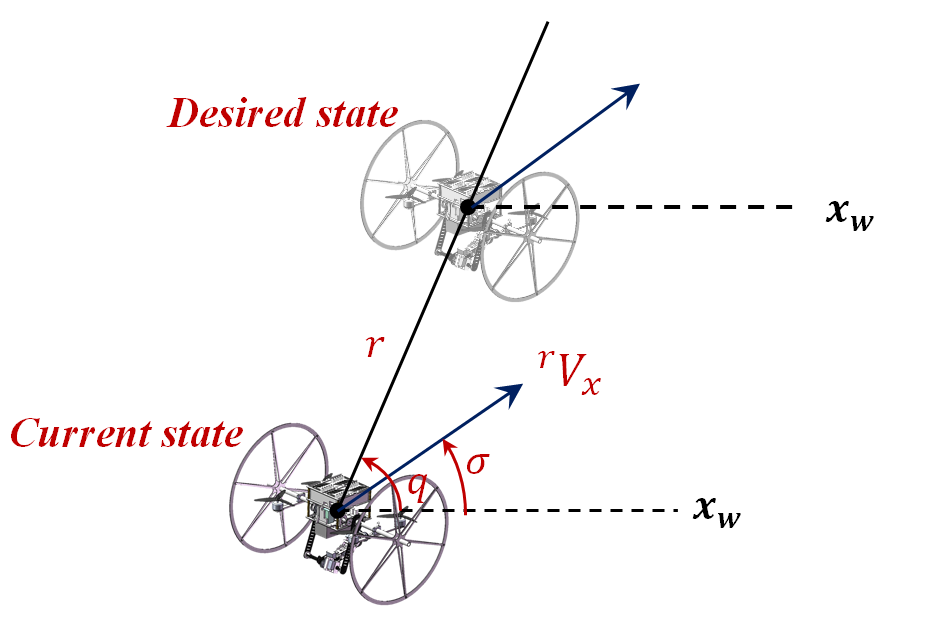}
	\caption{Illustration of yaw angular velocity control in Chat-PM.}
	\label{yaw_control}
\end{figure}

\textbf{Attitude Determination:} This module generates the desired angle for attitude control based on $\bm{F_d}$, $\psi$, and $^0\bm{f}_0$, with $\bm{F_d}$ representing the desired resultant force generated by the position controller. According to (1), the desired angle is calculated by
\begin{subequations}
\begin{equation}
\theta_d = atan \  {\frac{({F_{x,d}}-{{^0}f_{0,x}})cos{\psi} + ({F_{y,d}}-{{^0}f_{0,y}})sin{\psi}}{{-F_{z,d}}+mg-{{^0}f_{0,z}}}} 
\end{equation}
\begin{equation}
	\phi_d = atan \
	{\frac{-({F_{x,d}}-{{^0}f_{0,x}})sin{\psi} + ({F_{y,d}}-{{^0}f_{0,y}})cos{\psi}}{{-F_{z,d}}+mg-{{^0}f_{0,z}}}} 
\end{equation}
\end{subequations}
where $\theta_d$ and $\phi_d$ represent the desired pitch and roll angles, respectively, and $\psi_d$ is obtained by the common method for UAVs, see \cite{8968276,brescianini2013nonlinear}.
It is worth noting that the computed interference torque is ultimately included in $[u_2,u_3,u_4]$.
\subsection{Terrestrial-Mode Controller}
The terrestrial-mode controller is active when the Chat-PM operates/moves on surfaces with various inclinations, which is strikingly different from the aerial-mode controller, in the following distinctions
\begin{itemize}
\item There are kinematic constraints along $\bm{y}_r$ and $\bm{z}_r$ in terrestrial mode.

\item The yaw controller for the aerial mode can not accurately execute the desired motion command due to the resistance torque generated by gravity.

\item More thrust can be allocated to attitude control in terrestrial mode, owing to the support force.
\end{itemize}
The overall control scheme for terrestrial mode is shown in Fig. \ref{terrestrial-mode controller}.

\textbf{Position Control:} 
For terrestrial locomotion, the position controller should be designed with consideration of the kinematic constraints.
The controller receives the desired state ${{^w}P_d,{^w}V_d,{^w}A_d,\psi_d}$ given by the high-level planner, and outputs the desired acceleration along $\bm{x}_r$. The desired acceleration in the rolling frame ${{^r}A_{x,d}}$ can be obtained by
\begin{subequations}
\begin{equation}
{{^w}V_d'} = {K_{p,p}}({^w}P_e)+ {K_{d,p}}({^w}{\dot{P}}_e)+ {K_{i,p}}\int{{^w}P_e}+{^w}V_d	
\end{equation}	
\begin{equation}
{{^w}A_d'} = {K_{p,v}}({^w}V_e)+ {K_{d,v}}({^w}{\dot{V}}_e)+ {K_{i,v}}\int{{^w}V_e}+{^w}A_d   
\end{equation}
\begin{equation}
{^r}A_{x,d} = R{^r_w}({^w}A_d')
\end{equation}
\end{subequations}
where ${^w}P_e \triangleq {{^w}P_d} - {{^w}P}$, ${^w}V_e \triangleq {{^w}V_d'} - {{^w}V}$, and $K$ is the parameter of the controller.

\textbf{Yaw Control:} In previous studies \cite{9691888,8968276}, the yaw control is designed as the tangent function of the position deviation. Although the works have a good control performance on the ground, they are shown inefficient on surfaces with large angles, because the determined control parameters are difficult to adapt to surfaces with various inclinations.
In this letter, a proportional control law is utilized for the control of yaw angle. As conveyed in Fig. \ref{yaw_control}, the straight line connecting the current position and target position of Chat-PM is defined as Line of Sight (LOS), $q$ is the angle between LOS and $x_w$ axis, and $\sigma$ is the angle between the velocity vector of Chat-PM and $x_w$ axis. The control law of the yaw angle is shown below
\begin{equation}
\dot{\psi}  = \frac{d\sigma}{dt} =K\frac{dq}{dt}
\end{equation}
where $K$ is the proportional coefficient, regulating the angular rate changes of $\sigma$ and $q$. To improve the control performance, $K$ is designed to depend on the surface's inclination $\gamma$.
By performing fitting process to the data of steering experiments on surfaces with different angles, the obtained linear relation between $K$ and the inclination $\gamma$ is
\begin{equation}
 K = 0.137\gamma+1.1051
\end{equation}

\textbf{Thrust and Attitude Determination:} When Chat-PM is moving on the inclined surface, thrust $u_1$ and pitch angle $\theta$ are coupled, and due to the inequality constraint presented in the dynamics (12), multiple pairs of thrust and pitch can be solved to achieve specific acceleration. 
To achieve better operation capability, control stability, and energy efficiency of Chat-PM, we establish a second-order relationship between $\bm{F}$ and the surface's inclination $\gamma$ based on the experimental results of motion on the surfaces
\begin{equation}\lVert \bm{F} \rVert = (-0.002\gamma^2  +   0.0179\gamma+0.6728)mg\end{equation}
By solving the equations of (6)-(10), (12) and (18), the desired thrust and pitch can be uniquely determined, and the obtained solution for the pitch angle is
\begin{equation}
\theta_d=\operatorname{atan} \ \frac{m {^r}A_{x,d}+m g \cos \alpha \sin \gamma+u_r F_n+({ }^r f_0+f_w) \widehat{\boldsymbol{X}}_{\boldsymbol{r}}}{m g \cos \gamma+({ }^r f_0+f_w)\widehat{\boldsymbol{Z}}_{\boldsymbol{r}}-F_n}
\end{equation}
It should be noted that the attitude controller in terrestrial mode also follows the PID control strategy, but only the pitch angle $\theta_d$ is considered in the control objective, with the roll angle $\phi_d = 0 $ due to the contact between wheels and the surface.
\begin{figure*}[tbp]
	\centering
	\includegraphics[width=\linewidth]{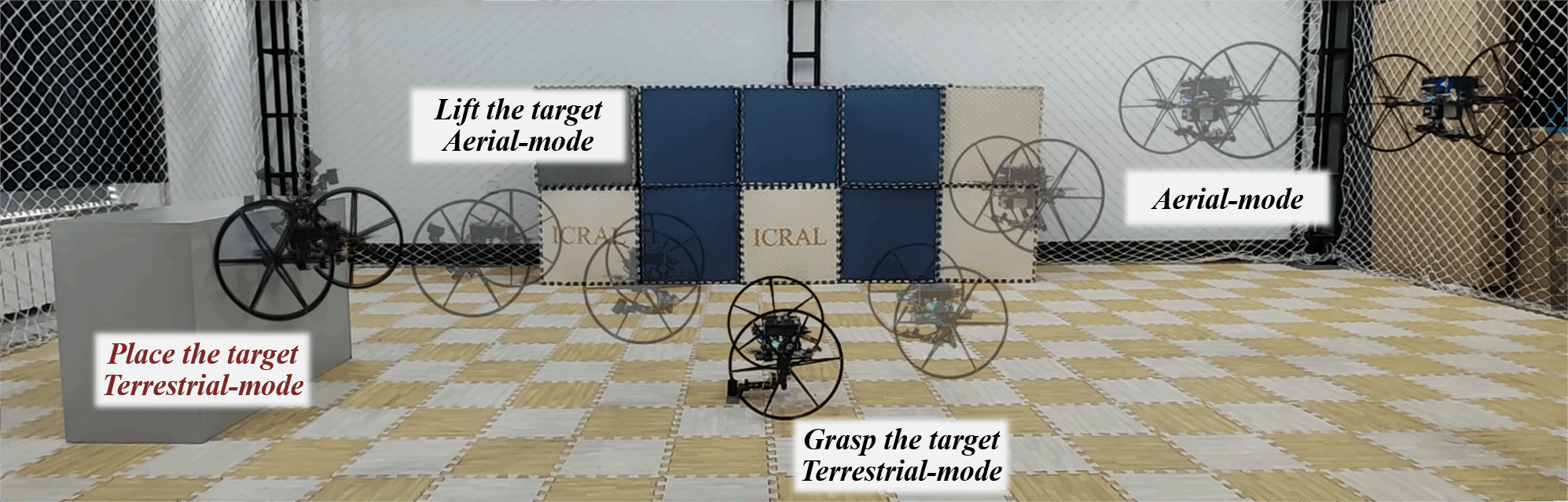}
	\caption{The motion trajectory of Chat-PM, which presents hybrid aerial/terrestrial locomotion and complex manipulation.}
	\label{aerial/terrestrial locomotion}
\end{figure*}
\subsection{Force Estimation}
The estimator developed in this subsection is used for providing the inclination $\gamma$ of the unknown surface that Chat-PM is in contact with, based on which the feedback gain can be scheduled according to (17) and the accurate pitch angle $\theta_d$ can be solved in terrestrial mode according to (19), thus better motion performance can be obtained for the movement on the inclined surface.
When Chat-PM contacts with the surface, it will be affected by the support force ${\bm{F_n}}$ along ${\hat{\bm{z}}}_{\bm{r}}$. Therefore, the inclination of the surface can be approximately obtained as the angle $\gamma$.
In this subsection, the external force estimator is developed via the methodology of moving horizon estimation (MHE). Since the reaction force from the robotic arm can be accurately modeled and the near-wall effect of Chat-PM is small enough to ignore, cf. \cite{2020Aerodynamic}, the estimated external force is almost equal to $\bm{F_n}$. As such, the estimation problem is formulated as the following discrete-time system
\begin{subequations}
	\begin{equation}
		\bm{x_{k+1}}=f\left(\bm{x}_k, \bm{u}_k\right)+\bm{w}_k
	\end{equation}
	\begin{equation}
		\bm{y}_k=h\left(\bm{x}_k, \bm{u}_k\right)+\bm{v}_k, k \geq 0
	\end{equation}
\end{subequations}
subject to the constraints
\begin{equation}
	\bm{x}_k \in {X}, \bm{u}_k \in {U}, \bm{w}_k \in {W}, \bm{v}_k \in {V}
\end{equation}
where $X, U, W, V$ are closed convex sets, $\bm{x}_k$ and $\bm{u_k}$ are the system state and control input at instant $k$, respectively. In order to estimate the external force, the state $\bm{x}_k$ in (21) is extended as
\begin{equation}
	\bm{x} = [P_x,P_y,P_z,V_x,V_y,V_z,{F}_{nx},{F}_{ny},{F}_{nz}]^T
\end{equation}
Note that the external forces are assumed to be constant within the estimation horizon. According to the differential flatness property \cite{2011Minimum}, the control input $\bm{u}_k$ is selected as $\bm{u} = [U_1,\phi,\theta,\psi]^T$,
$\bm{y}_k$ is the measurement output; the function $f(\cdot)$ can be obtained by discretizing the translational dynamics in (1); $h(\cdot)$ is the measurement function, given by $h(\cdot)=\left[P_x,P_y,P_z,V_x,V_y,V_z, U_1, \phi, \theta, \psi\right]^T$,
$\bm{w}_k$ and $\bm{v}_k$ denote the process noise and measurement noise, respectively, and they are assumed to be normally distributed sequences with zero mean and covariance of $Q$ and $R$, respectively.

To avoid the growth of the computational complexity over time, the estimation problem is considered within a limited horizon, with the estimation window receding with time. In this way, the state estimation is formulated as the following optimization problem
\begin{equation}
	\begin{gathered}
		\min _{\bm{x}_k, \bm{w}_k} \sum_{k=T-N}^T\left\|\bm{y}_k-h\left(\bm{x}_k, \bm{u}_k\right)\right\|_{R^{-1}}^2 
		+\sum_{k=T-N}^{T-1}\left\|\bm{x_{k+1}}-f\left(\bm{x}_k, \bm{u}_k\right)\right\|_{Q^{-1}}^2\\
		+\left\|\bm{\hat{x}_{T-N}}-\bm{\bar{x}_{T-N}}\right\|_{P_{T-N}^{-1}}^2
	\end{gathered}
\end{equation}
where $T$ is the current time; $N$ is the estimation horizon; the third term is the arrival cost, which approximately evaluates the estimation error beyond the estimation horizon; ${\bm{\hat{x}}}_{T-N}$ represents the state estimate at time $T-N$ calculated by MHE at time $T$; ${\bm{\bar{x}}}_{T-N}$ and $P_{T-N}$ express the priori state estimate and the covariance matrix of the priori error at time $T-N$, respectively, which are approximated by using forward dynamic programming and EKF without considering the constraints \cite{kocer2019aerial}. Finally, the external force estimate can be obtained via the online numerical solution of the optimization problem (23) subject to (21).

\section{Experiment Results}
In this section, the superiority of the configuration design and the effectiveness of the proposed control strategy for Chat-PM is demonstrated through a series of real-world experiments, including the manipulation tasks of grasping, carrying, placing and writing. More details about the experiments are presented in the attached video\footnote{https://youtu.be/zJkpJ1DHb4c}.         

\subsection{Grasping, Carrying and Placing Demonstration} 
\begin{figure}[tbp]
	\centering
	\includegraphics[width=0.5\textwidth]{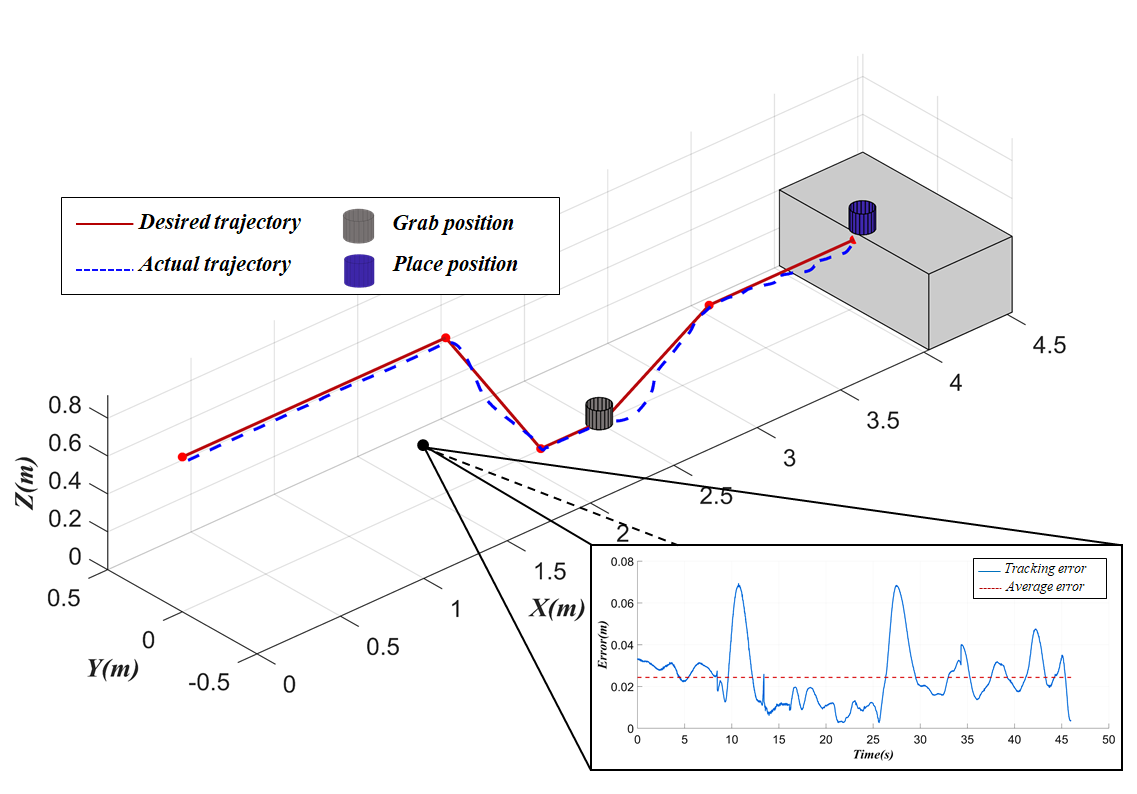}
	\caption{Tracking performance of both aerial mode and terrestrial mode controllers. The weight of the target object is $30g$.}
	\label{exp1-2}
\end{figure}

In this experiment, Chat-PM tracks a desired trajectory with hybrid aerial/terrestrial locomotion and executes the grasping-placing tasks, as shown in Fig. \ref{aerial/terrestrial locomotion}.
To be specific, Chat-PM is required to lift the object off the ground and place it on the box while fulfilling the specified velocity and acceleration constraints of $0.6 m/s$ and $0.2m/s^2$, respectively.
As shown in Fig. \ref{exp1-2}, 
Chat-PM performs an average tracking error of $2.25cm$ and a maximum tracking error of $6.52cm$ along the desired trajectory, and the maximum tracking error occurs in the ascent and descent stages.
Benefiting from the force and torque compensation via the 4-DOF arm model, Chat-PM is capable of maintaining an average tracking error of $2.84cm$ even in the presence of arm movement interference, which is comparable with the state-of-the-art result ($3.3cm$ in \cite{7989753}).
Besides, with the help of two passive wheels, Chat-PM can move along the wall with carrying the object to the top of the box. Meanwhile, by adopting the omnidirectional arm, Chat-PM can grasp and place the target in any direction.
To sum up, Chat-PM exhibits satisfactory trajectory-tracking performance and grasping-placing capability, which verifies the effectiveness of the proposed controllers and exhibits the potential of the manipulator.
\subsection{Precision of Manipulation ($\gamma = 90^o$)}
\begin{figure}[tbp]
	\centering  
	\subfigbottomskip=2pt 
	\subfigcapskip=-3pt 
	\subfigure[Circle data]{
		\label{Circle experiment}
		\includegraphics[width=0.46\linewidth]{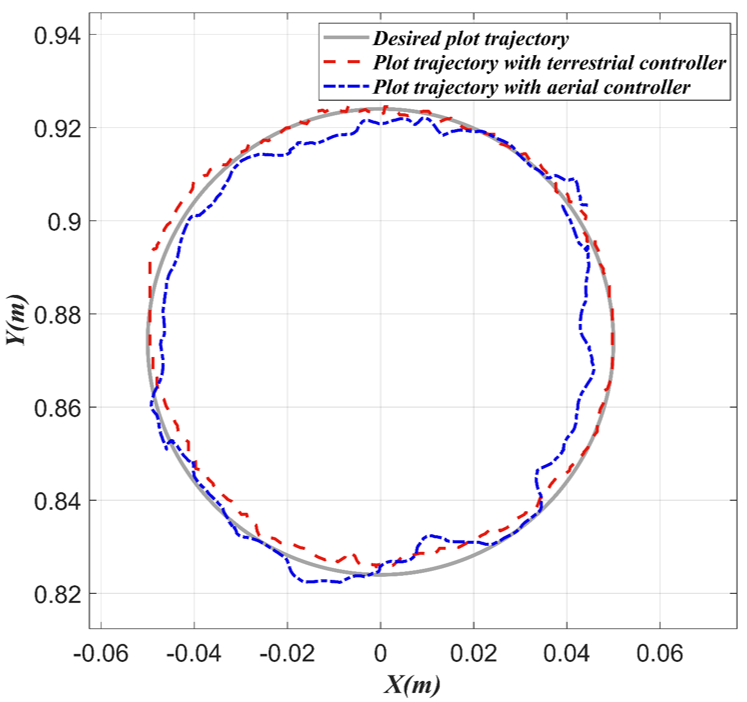}} 
	\subfigure[Letters data]{
		\label{Letters experiment}
		\includegraphics[width=0.47\linewidth]{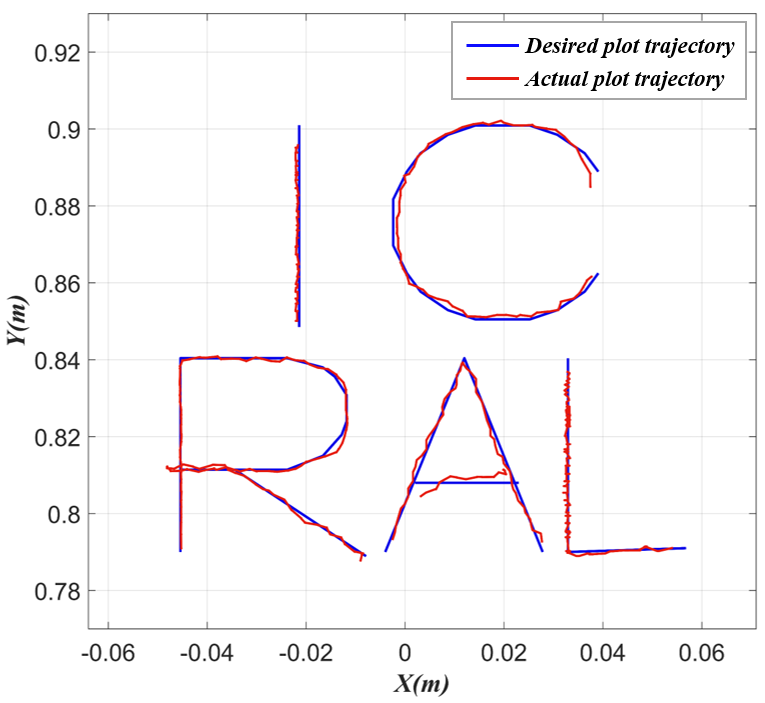}}
	\subfigure[Circle experiment]{
		\label{Circle experiment2}
		\includegraphics[width=0.47\linewidth]{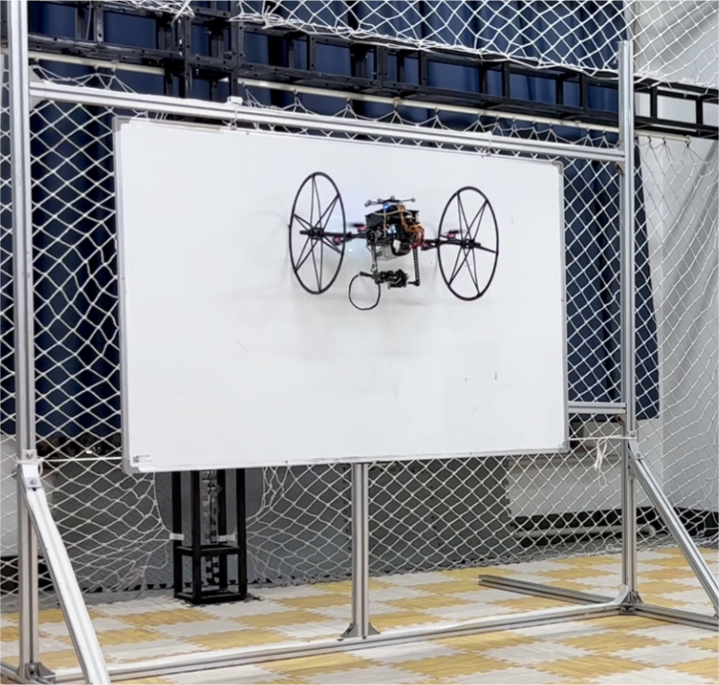}} 
	\subfigure[Letters experiment]{
		\label{Circle experiment2}
		\includegraphics[width=0.48\linewidth]{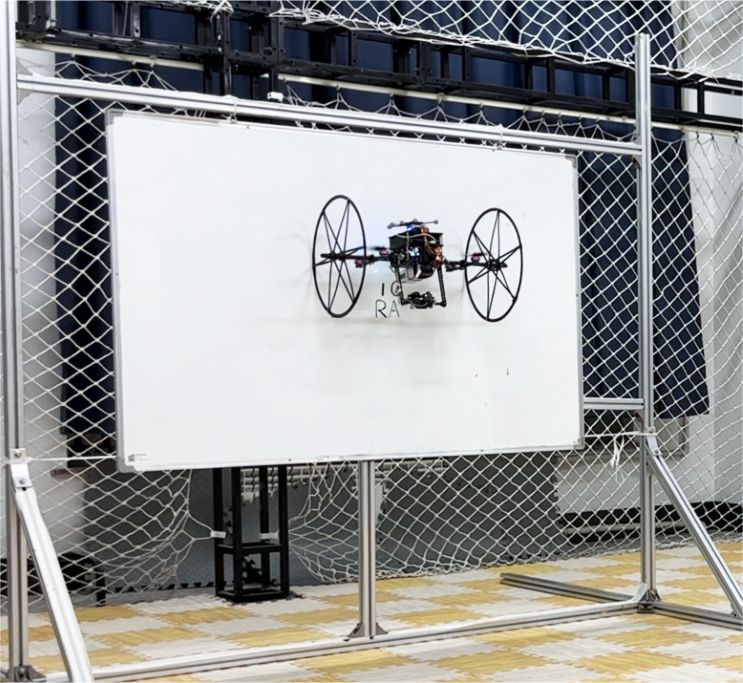}} 
	\caption{Manipulation performance of the manipulator in terrestrial mode }
	\label{exp2}
\end{figure}

To validate the manipulation precision of Chat-PM with aerial and terrestrial controllers, we conduct a series of experiments where Chat-PM draws on the whiteboard with $90^o$.
The first one, drawing the circle, is performed in both aerial and terrestrial modes, where Chat-PM executes the task with/without contacting with the whiteboard. The second one, writing the letters ``ICRAL", requires Chat-PM to switch frequently between aerial and terrestrial modes to achieve the precise draw of the five letters, which is more challenging.
As illustrated in Fig. \ref{Circle experiment}, the manipulation precision in the terrestrial mode is significantly better than the one in the aerial mode due to the kinematic constraints along $\bm{x_r}$ and $\bm{y_r}$ imposed by the support force.
Also, the manipulation performance of the proposed method is compared with the ones in the recent studies \cite{8928943,8401328}, as shown in Tab. \ref{exp2-table}.
The results indicate the Chat-PM in terrestrial mode achieves higher manipulation precision and the maximal error is only $0.52 cm$.
As shown in Fig. \ref{Letters experiment}, Chat-PM can track the letters ``ICRAL" trajectory with an average error of less than $0.831 cm$, which demonstrates the potential of the hybrid aerial/terrestrial structure in complex manipulation tasks. 
Besides, the comparison results show that the designed control strategy can effectively attach the surface and improve the manipulation precision. 
\subsection{Precision of Manipulation ($\gamma \neq 90^o$)}
\begin{figure}[htbp]
	\centering
	\includegraphics[width=0.5\textwidth]{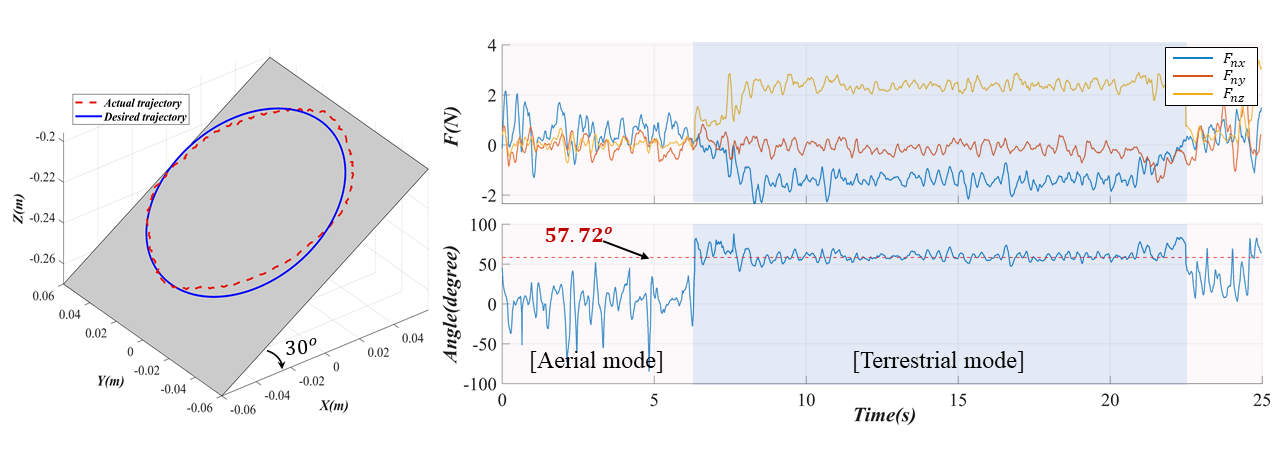}
	\caption{The operation trajectory and force estimation of MHE for the $30^o$ surface.}
	\label{exp3-1}
\end{figure}

\begin{figure}[htbp]
	\centering
	\includegraphics[width=0.5\textwidth]{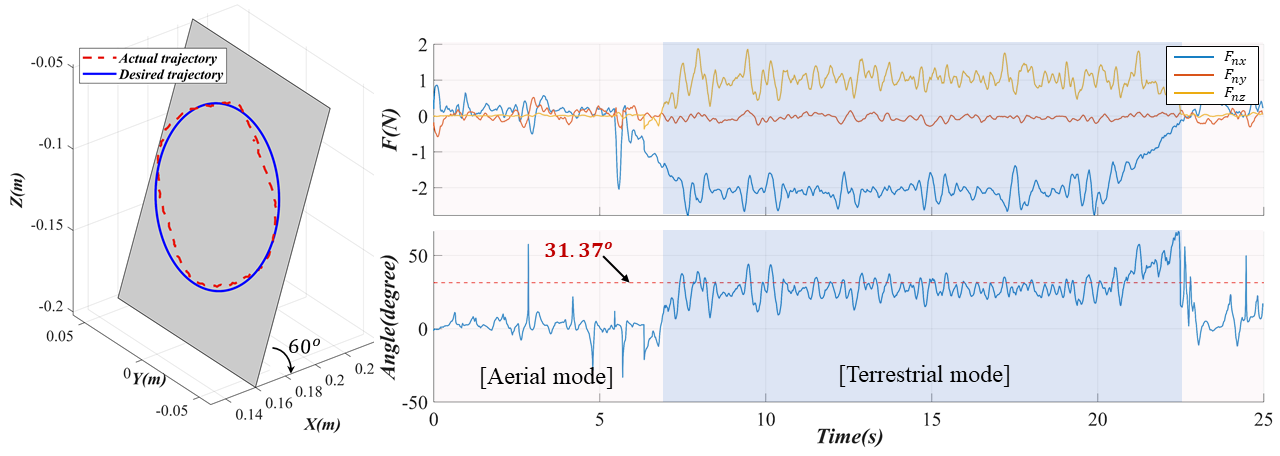}
	\caption{The operation trajectory and force estimation of MHE for the $60^o$ surface.}
	\label{exp3-2}
\end{figure}

The purpose of this experiment is to validate the manipulation precision of Chat-PM on unknown surfaces. An estimator is designed based on the MHE method to obtain the contact force and the inclination angle of the unknown surface, and the inclination angle $\gamma$ is used in the terrestrial controller to calculate a set of thrust $\bm{F}$ and pitch angle $\theta_d$ solution making Chat-PM move efficiently on the surface.
As shown in Fig. \ref{exp3-1} and Fig. \ref{exp3-2}, two circle experiments are conducted on surfaces with inclination angles of $60^o$ and $30^o$, respectively, and the angles between the horizontal direction and the contact forces estimated by MHE are $57.72^o$ and $31.37^o$, respectively.
Benefiting from the iterative optimization of estimation error via MHE method, the circle trajectory drawn by Chat-PM fits the desired circle curve with less errors, where the maximum errors are less than $2.2 cm$ and $1.3cm$ for the two inclination angles, respectively, showing that the MHE-based inclination angle estimation is precise.
Hence, the terrestrial model can be adjusted precisely, and thus Chat-PM is able to manipulate precisely on surfaces with unknown inclination angles.

\begin{table}[hbp]
	\caption{Comparison of manipulator trajectory tracking}
	\label{exp2-table}
	\vspace{-1em}
	\renewcommand{\arraystretch}{1.3}
	\begin{center}
		\begin{threeparttable}
			\begin{tabular}{c c c }
				\hline
				Method    & Maximal error($cm$) 	& Average error($cm$) 	 \\
				\hline
				Nava's \cite{8928943}    & $0.98$   				& $0.61	$	 \\
				Park's \cite{8401328}   & $3.04$   				& $-$ \\
				Aerial mode   & 1.24       		& 0.86		 \\
				Terrestrial mode  &\textbf{0.52}   	&\textbf{0.39}\\
				\hline
			\end{tabular}
		\end{threeparttable} 
	\end{center}
	\vspace{-1.3em}
\end{table}
\section{Conclusions}
This study investigates the design, modeling, control and force estimation for Chat-PM, a new hybrid aerial/terrestrial manipulator integrated by a quadrotor and a 4-DOF robotic arm, on different sloping surfaces (including the ground). 
The precision of the manipulator on inclined surfaces or walls can be significantly improved by the accurate modeling and control strategy.
A MHE-based estimator is designed for force estimation, allowing for autonomous manipulation on surfaces with unknown inclination angles. Several experiments are conducted to demonstrate the effectiveness of the proposed control strategy and the superiority of the Chat-PM in improving manipulation precision in terrestrial mode. Future attention will be paid to the planning problems for Chat-PM, including unified aerial-terrestrial motion planning and unified vehicle-arm planning. 

\addtolength{\textheight}{-12cm}   




\bibliographystyle{IEEEtran}      

\bibliography{ref}                        

\end{document}